\documentclass{article}
\PassOptionsToPackage{numbers, compress}{natbib}
\usepackage[preprint]{edited_neurips_2024}

\usepackage{graphicx}
\usepackage{framed,multirow,multicol}
\usepackage{amsmath,amssymb,amsfonts}
\usepackage{booktabs}
\usepackage{xurl}
\usepackage{dblfloatfix}               
\usepackage[dvipsnames]{xcolor}
\usepackage{colortbl}
\usepackage[hidelinks]{hyperref}       
\usepackage{titletoc}                  
\usepackage{bbm}
\usepackage{soul}                      
\usepackage{float}                     

\newtheorem{theorem}{Theorem}[section]

\newtheorem{definition}{Definition}[section]

\newcommand{\shadedText}[1]{
\noindent\colorbox{teal!5}{
  \parbox{\dimexpr\linewidth-4\fboxsep}{
      #1
    }
  }
}

\usepackage{tikz}
\definecolor{YaleBlue}{rgb}{0.059,0.302,0.573}
\definecolor{forestgreen}{rgb}{0.133,0.549,0.133}
\definecolor{crimson}{rgb}{0.863,0.078,0.235}

\makeatletter
\newcommand{\linebreakand}{%
  \end{@IEEEauthorhalign}
  \hfill\mbox{}\par
  \mbox{}\hfill\begin{@IEEEauthorhalign}
}
\makeatother

\title{DiffKillR: Killing and Recreating Diffeomorphisms for Cell Annotation in Dense Microscopy Images}

\author{
\textbf{Chen Liu}$^{1 *}$ \quad
\textbf{Danqi Liao}$^{1 *}$ \quad
\textbf{Alejandro Parada-Mayorga}$^{2 *}$ \and
\textbf{Alejandro Ribeiro}$^{3}$ \quad
\textbf{Marcello DiStasio}$^{1}$ \quad
\textbf{Smita Krishnaswamy}$^{1}$ \vspace{6pt}\\
{\small $^1$Yale University \quad $^2$University of Colorado Denver \quad $^3$University of Pennsylvania \vspace{6pt}}\\
{\small *~These authors are joint first authors.}\\
{\small Please direct correspondence to: \url{smita.krishnaswamy@yale.edu}.}
}

\begin{document}
\maketitle

\begin{abstract}

The proliferation of digital microscopy images, driven by advances in automated whole slide scanning, presents significant opportunities for biomedical research and clinical diagnostics. However, accurately annotating densely packed information in these images remains a major challenge. To address this, we introduce DiffKillR, a novel framework that reframes cell annotation as the combination of archetype matching and image registration tasks. DiffKillR employs two complementary neural networks: one that learns a diffeomorphism-invariant feature space for robust cell matching and another that computes the precise warping field between cells for annotation mapping. Using a small set of annotated archetypes, DiffKillR efficiently propagates annotations across large microscopy images, reducing the need for extensive manual labeling. More importantly, it is suitable for any type of pixel-level annotation. We will discuss the theoretical properties of DiffKillR and validate it on three microscopy tasks, demonstrating its advantages over existing supervised, semi-supervised, and unsupervised methods. The code is available at \url{https://github.com/KrishnaswamyLab/DiffKillR}.

\end{abstract}

\section{Introduction}

Rapid advances in microscopy technologies, including whole slide imaging for clinical diagnostics, have dramatically increased the availability of digital tissue images~\cite{histology_staining1}. Medical centers around the world now routinely employ whole slide scanning, generating an ever-growing repository of highly detailed and feature-rich images~\cite{farahani2015whole}. However, fully leveraging this vast resource is challenging due to the complexity of identifying, segmenting, and extracting specific features from densely packed tissue elements~\cite{morales2021artificial}. These features include cellular polarity, orientation, subcellular structures, and specialized cytostructural components such as the extracellular matrix, pathological protein aggregates, and foreign materials~\cite{krithiga2021breast, li2022comprehensive}. The immense volume and intricate details of these images make exhaustive manual annotation impractical. This challenge underscores the urgent need for label-efficient methods that can perform these critical tasks with minimal human intervention~\cite{yu2021accurate, wang2022weakly, jiang2022weakly, huang2023push, li2024generating, sun2024logit, zhang2024provably}.

Current approaches for cell annotation primarily rely on supervised learning methods that map images to desired outputs using dense labeling~\cite{UNet, nnUNet, MedT}. Unsupervised and semi-supervised methods, on the other hand, aim to uncover patterns in local or global image features~\cite{CUTS, kochetov2024unseg, feng2018semi, liao2024assessing, he2020semi}. Despite their potential, these approaches often do not fully exploit the unique characteristics of microscopy images~\cite{elyan2022computer}. Cells exhibit diverse shapes, poses, and morphometric features, but notably, a small set of archetypes can represent most cells~\cite{venkat2024aanet, rusin2023evolution}. This observation led us to explore a novel approach of transforming the cell annotation task into an image registration task. \sethlcolor{Peach!20}\hl{When two cells are sufficiently similar, differing only by a diffeomorphism, we can compute the warping field between them. This enables a pixel-perfect mapping of annotation from one cell to the other.} \sethlcolor{YellowGreen!30}\hl{To ensure that the computation of diffeomorphisms is robust, we first need to devise a measure of similarity between cells that is invariant to differences correctable by a diffeomorphism.}

Based on this insight, we developed \textbf{DiffKillR}, a novel framework that ``\textbf{kill}s'' and ``\textbf{r}ecreates'' \textbf{diff}eomorphisms to facilitate label-efficient annotation. Starting with expert annotations for a small set of archetypal cells, DiffKillR propagates these annotations to all cells of interest using two complementary neural networks. The first network, the ``killing'' network termed \sethlcolor{YellowGreen!30}\hl{DiffeoInvariantNet}, is agnostic to diffeomorphisms and learns a specialized representation space for cells, allowing us to match cells based on diffeomorphism-invariant features. The second network, the ``recreating'' network termed \sethlcolor{Peach!20}\hl{DiffeoMappingNet}, is sensitive to diffeomorphisms and learns to map any new cell to the matching archetypal cell. The diffeomorphism between them can then be applied directly to any pixel-level annotation. This two-stage framework effectively enables label propagation for all cells using a selective set of annotated archetypes. Our main contributions include the following.

\begin{itemize}
    \item Introducing DiffKillR, a novel framework that reframes cell annotation as the combination of archetype matching and image registration tasks.
    \item Designing two complementary neural networks that are respectively invariant and sensitive to diffeomorphisms, for label-efficient cell annotation.
    \item Unit testing the effectiveness of the networks on matching cells and mapping diffeomorphisms, respectively.
    \item Validating the performance of DiffKillR on microscopy datasets on diverse tasks, including cell counting, cell orientation prediction, and few-shot cell segmentation.
\end{itemize}

\section{Related Works}

In this section, we review existing work on cell segmentation and the application of diffeomorphisms in medical image analysis. Although our approach is not confined to cell segmentation and can be extended to various microscopy tasks, cell segmentation is one of the most widely studied fields, with an abundance of relevant literature.

\subsection{Cell Segmentation}

\paragraph{Traditional methods for cell segmentation}
Automatic cell segmentation is a crucial step in various biological image analysis tasks. Traditionally, image processing techniques such as line/edge detection~\cite{tradseg_edge}, graph cuts~\cite{tradseg_graph_cut}, active contours~\cite{tradseg_active_countour}, watershed~\cite{watershed}, and level set~\cite{tradseg_level_set} have been employed for segmentation. However, these methods struggle with the high variance of microscopy images due to overlapping or touching cells, uneven illumination, and complex cell morphologies.

\paragraph{Supervised learning for cell segmentation}
Deep learning has emerged as a powerful alternative for cell segmentation, offering superior performance and the ability to learn complex patterns from large datasets~\cite{DL_medseg}. Supervised learning has been widely used for semantic segmentation, where the goal is to assign each pixel in the image to a specific cell class. This has led to significant advances in cell segmentation accuracy. U-Net~\cite{UNet}, a general purpose deep learning framework and a milestone in semantic segmentation, inspired many works to adopt a neural network with an encoder and a decoder with skip connections for cell segmentation~\cite{SwinUNet, zhu2022segmentation, Omnipose, Cellpose}.

\paragraph{Self-supervised and weakly-supervised cell segmentation}
Self and weak supervision are explored to overcome the dependency on a substantial amount of labeled training data~\cite{CUTS}. PSM~\cite{PSM} aggregated gradient information from a network after self-supervised training to generate self-activation maps that serve as pseudo-labels for training. WSISPDR~\cite{WSISPDR} combined a cell location prediction network with a backpropagation technique to refine these predictions and guide graph-cut-based segmentation of individual cells. LACSS~\cite{LACSS, LACSS2} used a location proposal network to predict potential cell locations and a segmentation network to refine those locations into precise cell segmentations. 

\paragraph{Instance segmentation techniques}
For detailed analysis, it is often necessary to identify and distinguish individual cell instances. This requires instance segmentation, which goes beyond classifying pixels and aims to localize and delineate the boundaries of each cell in the image. One popular approach is to apply a post-processing step, such as watershed~\cite{watershed}, on the semantic segmentation mask to break it into masks of individual cells~\cite{cellseg2d, DeepSea}. Some other methods predict an intermediate pseudo-image, usually a ``distance map'', and convert it to instance-level masks~\cite{Seg_distance_map, MicroNet}. Alternatively, one can follow the segmentation-by-detection approach as introduced by Mask R-CNN~\cite{mask_rcnn} and extend it to cell instance segmentation~\cite{cellseg_mask_rcnn}.

As will be discussed later, our proposed method is intrinsically instance-level with no additional post-processing required. Moreover, it can be applied to any pixel-level annotation, including but not limited to binary segmentation masks, which are the most frequently investigated.

\subsection{Diffeomorphism in Medical Image Analysis}

\paragraph{Image registration}
A diffeomorphism is defined as ``a map between manifolds which is differentiable and has a differentiable inverse"~\cite{mathworld_diffeomorphism}. Diffeomorphisms have played a crucial role in image registration, a fundamental task in medical image analysis~\cite{SymmetricDiffeo}. Image registration aims to align images by estimating a diffeomorphism between them~\cite{LargeDeformationDiffeo, he2022quantifying, jiang2021weakly}. The estimated diffeomorphism is usually represented as a vector field $\mathcal{W} \in \mathbb{R}^{H \times W \times 2}$ for an image $x \in \mathbb{R}^{H \times W \times C}$ and defines the direction and distance that each point in the image should move to align with the target image. Estimation is guided by a metric, typically mean squared error or normalized cross-correlation~\cite{Matching_NCC}. Popular optimization techniques include ANTS~\cite{ANTS}, NiftyReg~\cite{NiftyReg}, or neural networks~\cite{Voxelmorph, RecursiveRefinement}.

\paragraph{Atlas-based image segmentation}
In the domain of volumetric images, particularly magnetic resonance imaging~(MRI), researchers have proposed image segmentation through registration to a common atlas~\cite{MedSeg_Registration}. After registering each image to a pre-defined atlas using a diffeomorphism, the segmentation information from the atlas can be warped and transferred back to each image~\cite{MedSeg_Registration, LTNet}. However, it is highly dependent on the quality of the atlas and the alignment with individual images. Historically, this method has been used mainly in volumetric images of anatomic structures such as brains and major organs that show little to moderate morphological variance between subjects. Due to the morphological diversity of cells~\cite{Morphological_diversity}, it is difficult to define a common atlas or produce a high-quality alignment towards the atlas. As a result, to the best of our knowledge, this method has not been adopted in the field of microscopy images.

Our proposed method extends this approach to cell microscopy images by overcoming the challenge of a large variance in morphology among different cells. Through a cell archetype matching process in the latent space, we reduce the morphology variance and registration difficulty between the matched cells and enable the adoption of this powerful segmentation and label transfer technique.

\section{Methods}

\begin{figure}[!tb]
\centering
\includegraphics[width=\textwidth]{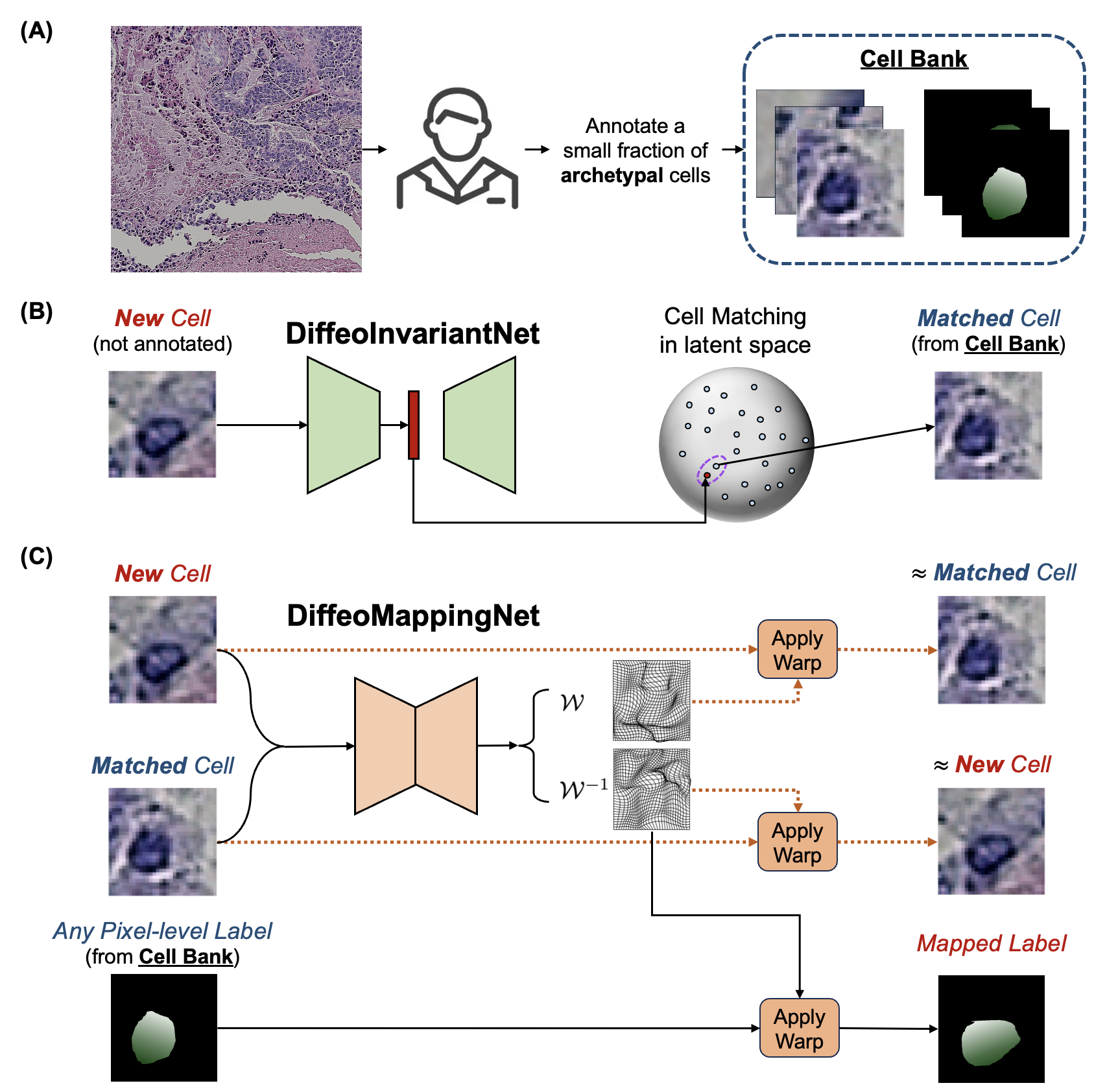}
\caption{Proposed DiffKillR framework. (\textbf{A)} A small set of annotated cells forms a cell bank. \textbf{(B)} DiffeoInvariantNet learns a latent space that is invariant to common diffeomorphisms. For each new cell, it finds the closest archetypal cell within the cell bank. \textbf{(C)} DiffeoMappingNet transforms the label to the new cell using the pairwise diffeomorphism computed via image registration. }
\label{fig:workflow}
\end{figure}

Our framework (Figure~\ref{fig:workflow}) consists of two stages: matching new cells with annotated archetypal cells by encoding them in a learned latent space that is invariant to augmentations (Section~\ref{sec:diffeoinvariantnet}), and computing the forward and inverse diffeomorphic warping fields for each matched pair (Section \ref{sec:diffeomappingnet}). The forward field maps new cells to their matched archetypes, while the inverse field reverses this mapping. During inference, the learned inverse field is applied to the annotations of the matched cells, mapping the annotations to the new cells.

\subsection{DiffeoInvariantNet: learning diffeomorphism invariance}
\label{sec:diffeoinvariantnet}
DiffeoInvariantNet is an autoencoder that matches new cells to annotated archetypal cells. It is specifically trained to ignore (that is, ``kill'') diffeomorphisms in cells, and thus we can perform cell matching based on proximity in its diffeomorphism-invariant latent space~(Figure~\ref{fig:workflow}(B)).

\subsubsection{Data augmentation with realistic diffeomorphisms}
\label{sec:realistic_diffeo}
\begin{figure}[!tb]
\centering
\includegraphics[width=\textwidth]{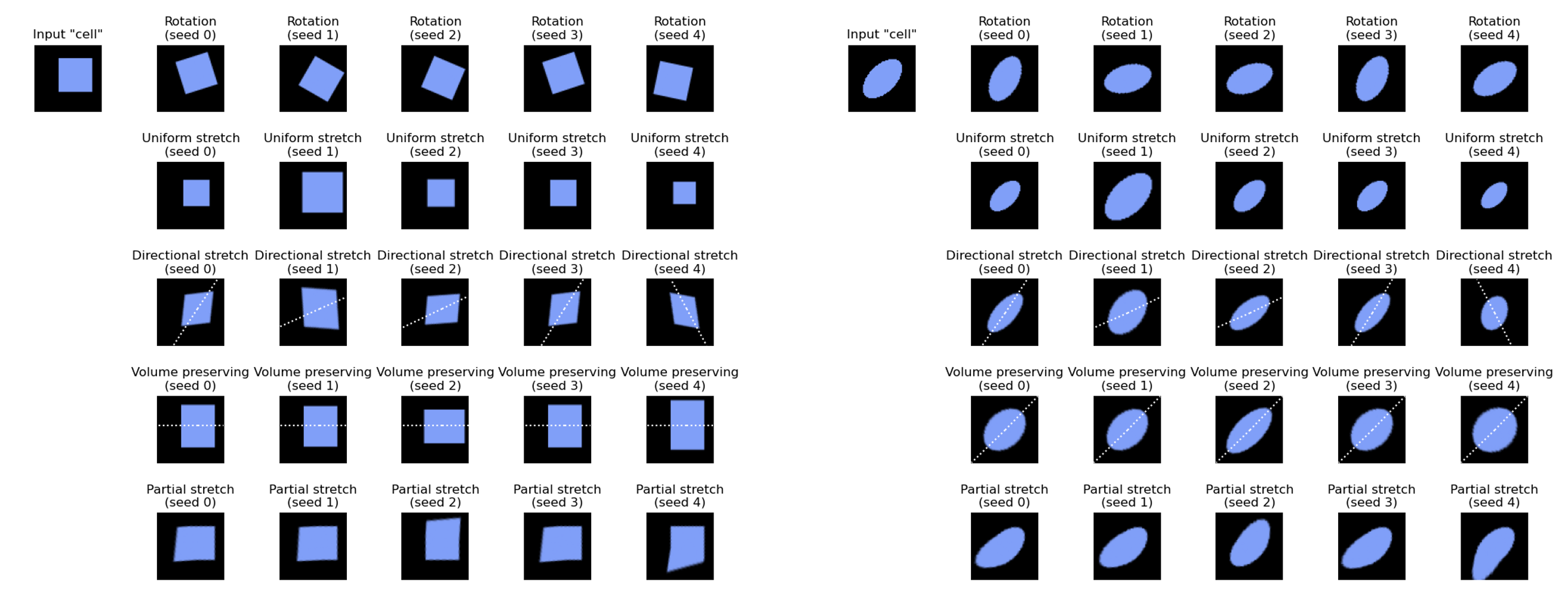}
\caption{Realistic diffeomorphisms, illustrated using a square and an ellipse. }
\label{fig:realistic_diffeomorphisms}
\end{figure}

We incorporate a set of realistic diffeomorphisms commonly observed in microscopy images, including rotation, uniform stretch across all directions, directional stretch along a random direction, volume-preserving stretch that stretches along one axis and shrinks the orthogonal axis, and partial stretch where only part of the cell is stretched and the rest remains intact. These realistic diffeomorphisms are illustrated in Figure~\ref{fig:realistic_diffeomorphisms}.

\subsubsection{``Killing'' diffeomorphisms}
To ``kill'' diffeomorphisms, we optimize the latent space of DiffeoInvariantNet to embed similar cell patches to the same location. Here, similar is defined as cell patches that are diffeomorphic to each other, using the data augmentation above. DiffeoInvariantNet becomes a diffeomorphism-invariant autoencoder, after joint optimization by a latent space contrastive loss~\cite{SimCLR} and an image reconstruction loss, as described in Eqn~\eqref{eqn:loss_diffeoinvariantnet}.

\scalebox{0.9}{
\begin{minipage}{1.1\linewidth}
\begin{equation}
\label{eqn:loss_diffeoinvariantnet}
\mathcal{L}_{\textrm{ DiffeoInvariantNet}} =
 \overset{\text{image reconstruction}}{\overbrace{
 \frac{1}{N}\sum_{i=1}^N ||x_i - \hat{x}_i||^2 
 }} +
 \overset{\text{latent contrastive}}{\overbrace{ 
 \frac{1}{2N}\sum_{i=1}^{2N} \sum_{\substack{j = 1 \\ j \neq i}}^{2N}  -\log \frac{\exp(\text{sim}(\mathbf{z}_i, \mathbf{z}_j) / \tau)}{\sum_{k=1}^{2N} \mathbbm{1}_{[k \neq i]} \exp(\text{sim}(\mathbf{z}_i, \mathbf{z}_k) / \tau)}
 }}
\end{equation}
\end{minipage}
}

Here, $z_i, z_j$ are the corresponding embeddings of $x_i, x_j$ in the latent space. $\text{sim}(\mathbf{u}, \mathbf{v}) = \frac{\mathbf{u}^\top \mathbf{v}}{||\mathbf{u}||\cdot ||\mathbf{v}||}$ is the cosine similarity between a pair of embeddings. $\tau$ is the temperature parameter that scales the similarity. $N$ is the batch size.

\subsection{DiffeoMappingNet: mapping pairwise diffeomorphisms}
\label{sec:diffeomappingnet}

DiffeoMappingNet computes the pairwise diffeomorphism between any new cell and the corresponding archtypal cell~(Figure~\ref{fig:workflow}(C)). Practically, it is an image registration network, and can be drop-in replaced by any latest neural network architecture for diffeomorphic image registration. In our implementation, we offer the flexibility to choose from UNet~\cite{UNet}, VM~\cite{VM}, VM-Diff~\cite{VM_diff}, and CorrMLP~\cite{CorrMLP}.

\subsubsection{Learning the pairwise diffeomorphisms}

The objective of image registration is to predict a warping field that spatially aligns a moving image with a fixed image. The warping field $\mathcal{W} \in \mathbb{R}^{H \times W \times 2}$ specifies the pixel-level transformations needed to match the moving image to the fixed one. DiffeoMappingNet is trained to produce pairs of warping fields \(\mathcal{W}\) and $\mathcal{W}^{-1}$ for cells and their purterbed verisons via common diffeomorphisms. During inference, the model first registers each new cell with the closest match from the cell bank and then uses the computed inverse field $\mathcal{W}^{-1}$ to map the annotations from the annotated archetypal cell to the new cell. This process can be parallelized for efficiency.

DiffeoMappingNet is optimized by an alignment loss that minimizes the L2 distance between the registered image and the fixed image, along with a cycle consistency loss to ensure reversibility of the transformation, as described in Eqn~\eqref{eqn:loss_diffeomappingnet}.

\begin{equation}
\label{eqn:loss_diffeomappingnet}
\mathcal{L}_{\textrm{ DiffeoMappingNet}} =
 \overset{\text{alignment}}{\overbrace{
 \sum_{i=1}^N || x_i - (\mathcal{W} \circ \tilde{x}_i) ||^2
 }} +
 \overset{\text{cycle consistency}}{\overbrace{ 
 \sum_{i=1}^N || \tilde{x}_i - \left( \mathcal{W}^{-1} \circ (\mathcal{W} \circ \tilde{x}_i) \right) ||^2
 }}
\end{equation}

Here, $x_i$ is an archetypal cell in the cell bank, and $\tilde{x}_i$ is its perturbed version transformed by realistic diffeomorphisms as explained in Section~\ref{sec:realistic_diffeo}. The notation $\mathcal{W} \circ x$ denotes applying the warping field $\mathcal{W}$ onto the image $x$.

\subsubsection{``Recreating'' diffeomorphisms and mapping annotations}

During inference, DiffeoMappingNet predicts the warping fields $\mathcal{W}$ and $\mathcal{W}^{-1}$ between the new cell and its best match from the cell bank. Then, we apply the warping to the ground truth label, and the result will be the label for the new cell. It is essential to note that this framework can map any pixel-level label, including but not limited to segmentation masks. If the user provides archetypal annotations of subcellular structures, directional axes, or any other labels, applying the warping field will map them in the same way.

\section{Theoretical Results}
In this section, we discuss the mathematical properties of our proposed DiffKillR framework, especially DiffeoInvariantNet. We show that any bandlimited continuous deformation on cell patches can be uniquely determined by a finite set of deformations, realized by cell patch augmentation. We also derive an upper bound on the error of cell matching using DiffeoInvariantNet.

\subsection{Diffeomorphism in the lens of Algebraic Signal Processing}

Let $G$ be a diffeomorphism group from $\mathbb{R}^2$ to $\mathbb{R}^2$, and let $L_{2}(G)$ be the set of continuous and square-integrable functions on $G$. 

Given $\boldsymbol{a}\in L_{2}(G)$, we can quantify the effect of a continuous deformation on the image signal $f:\mathbb{R}^{2}\to \mathbb{R}$ by the following equation:

\begin{equation}\label{equ_rho_ideal}
\rho\left(
         \boldsymbol{a}
     \right)
          \left\lbrace 
                f 
          \right\rbrace
          =
          \int_G 
               \boldsymbol{a}(g)
               \mathbf{T}_g 
                   fd\mu(g),
\end{equation}

where $\mathbf{T}_g : \mathbb{R}^{2} \to \mathbb{R}^{2}$ represents the group transformation $g\in G$ on $\mathbb{R}^2$.

Intuitively, function $\boldsymbol{a}$ assigns a weight to every deformation in $G$ and $\rho( \boldsymbol{a})\left\lbrace f \right\rbrace $ can be seen as an averaged version of $f$ when weighted by $T_g$. There exists a $ \boldsymbol{a}^{\star}$ such that $\rho( \boldsymbol{a}^{\star})\left\lbrace f \right\rbrace $ reproduces $f$. Therefore, to learn any continuous deformation in $G$, we need to learn the ideal $\boldsymbol{a}^{\star}$.

\subsection{Covers for diffeomorphism group and bandlimited deformations}
In this subsection, we show that the infinite dimensional transformation $\boldsymbol{a}$ can be characterized by a finite number of its
realizations. Such a number depends on the complexity of $\boldsymbol{a}$. We leverage the notion of
Fourier decomposition under the light of algebraic signal processing. 


\vspace{8pt}
\noindent
\scalebox{0.92}{
\begin{minipage}{1.08\linewidth}
\begin{definition}
\label{def_band_limited_filters_L1G}
Let $\boldsymbol{S}: L_{2}(G) \to L_{2}(G)$ be a positive definite self-adjoint operator with spectrum in $[0,\infty)$ and let $\int H(\lambda)dm(\lambda)$ be the decomposition of $L_2(G)$ as a direct integral indexed by the spectrum of $\boldsymbol{S}$. Let $F: L_2(G)\to \int H(\lambda)dm(\lambda)$ be a projection operator. Then, if the action of $F$ on $\boldsymbol{a}\in L_2(G)$ has support on $[0, \omega]$, the deformation $\boldsymbol{a}$ is $\omega$-bandlimitted. The set of all $\omega$-bandlimitted deformations is represented by $\mathcal{PW}_{\omega}(\boldsymbol{S})$.
\end{definition}
\end{minipage}
}
\vspace{4pt}

The value of $\omega$ in Definition~\ref{def_band_limited_filters_L1G} naturally provides a complexity measure of the deformations in $\mathcal{PW}_{\omega}(S)$. Deformations with fast variations and high complexity are associated with large $\omega$, while low variation and low complexity ones are associated with small $\omega$.

Furthermore, we introduce the definition for the minimal covering radius $\epsilon^{\star}(\widehat{G})$, the smallest radius required for a cover based on a finite subset $\widehat{G}$ in $G$.

\vspace{8pt}
\noindent
\scalebox{0.95}{
\begin{minipage}{1.04\linewidth}
\begin{definition}
\label{smallest_cover_G}
Let $B(g, \epsilon)$ be the open ball of geodesic radius $\epsilon$ centered at $g\in G$ and let $\widehat{G}$ be a finite subset of $G$. Let $Y_{\widehat{G}}(\epsilon):= \bigcup_{g\in \widehat{G}}B(g,\epsilon)$ and let $\epsilon_\textrm{max}(G):= \max_{g,g' \in G} d_{geo}(g,g^{'}),$ where $d_{geo}(\cdot, \cdot)$ is the geodesic distance on $G$. If $\epsilon_\textrm{max}(G)<\infty$, there exists $\epsilon(\widehat{G})>0$ such that $Y_{\widehat{G}}\left( \epsilon(\widehat{G}) \right)$ covers $G$ for any finite subset $\widehat{G}\subset G$. We denote by $\epsilon^{\star}(\widehat{G})$ the smallest $\epsilon(\widehat{G})\in (0, \epsilon_\textrm{max}(G)]$ such that $Y_{\widehat{G}}\left(\epsilon(\widehat{G})\right)$ covers $G$.
\end{definition}
\end{minipage}
}
\vspace{4pt}

Now, we can state the following theorem.

\vspace{4pt}
\shadedText{
\begin{theorem}
\cite[Adapted from Theorem~1.6]{pesenson2000sampling}
\label{thm:pesenson} 
Let $G$ be a Lie group and $\widehat{G}$ a finite subset of $G$. Then there exists a constant $C_{0}>0$ such that every deformation in $\mathcal{PW}_{\omega}(\boldsymbol{S})$ is uniquely determined by its values on $\widehat{G}$ as long as
\begin{equation} \label{equ:thm1_ineqality}
\epsilon^{\star}(\widehat{G})
          < 
 (C_0\omega)^{-1} 
       \leq 
       \epsilon_\textrm{max}(G) 
.
\end{equation}
\end{theorem}
}

We can see that every $\omega$-bandlimited deformation can be uniquely determined by some combinations of elements in $\widehat{G}$ since there exists a constant $C_{0}$ that satisfies Equation \ref{equ:thm1_ineqality}.

\subsection{Error bound on cell matching}
\label{sec_matching_operator}
In this subsection, we derive the error bound on the matching operator $\boldsymbol{M}$ that matches new cell image $\widehat{s}$ with reference cells in cell bank $\mathbf{T}_{g_i}\{ s_{j} \}$ for $i=1,2,\ldots, m$ and $j=1,2,\ldots, n$. $m$ denotes the number of augmentations performed on each annotated cell patch $s_j$.

The new cell $\widehat{s}$ is affected by an unknown deformation $\mathbf{T}_{g}$ in $G$.
Encoder $\mathbf{\Phi}$ of the DiffeoInvariantNet outputs the embeddings of the input cells. Then, for any element in $\text{Range}(\mathbf{\Phi})$, the matching operator $\boldsymbol{M}$ matches the test cell with the closest reference cell in embedding space: 

\scalebox{0.85}{
\begin{minipage}{1.1\linewidth}
\begin{equation}
\label{eq_matching_operator}
 \boldsymbol{M}
   \left\lbrace 
          \mathbf{\Phi}\left\lbrace \widehat{s} \right\rbrace
   \right\rbrace
   =
    \arg\min_{i,j} \left\Vert 
      \mathbf{\Phi}
            \left\lbrace 
                  \mathbf{T}_{g_{i}}
                         \left\lbrace 
                               s_j
                         \right\rbrace      
            \right\rbrace
      -
      \mathbf{\Phi} 
        \left\lbrace 
           \widehat{s}
        \right\rbrace
\right\Vert
\end{equation}    
\end{minipage}
}

\vspace{4pt}

\shadedText{
\begin{theorem}\label{theorem:error_bound}
Let $\boldsymbol{M}$ be the matching operator and $\mathbf{T}_{g_i} \left\lbrace s_j \right\rbrace =
\boldsymbol{M} \left\lbrace \mathbf{\Phi}\left\lbrace \widehat{s} \right\rbrace \right\rbrace$ for the test deformed cell $\widehat{s}$. If $Y_{\widehat{G}}(\epsilon) =  \bigcup_{g_{i}\in\widehat{G}}B(g_{i},\epsilon)$ is the minimum covering of $G$ and $\mathbf{\Phi}$ is $L$-Lipschitz, then it follows that 
\begin{equation*}
\Vert \mathbf{\Phi} \left\lbrace \mathbf{T}_{g_{i}} \left\lbrace s_j \right\rbrace \right\rbrace
- \mathbf{\Phi} \left\lbrace  \widehat{s} \right\rbrace \Vert
\leq
L \epsilon \Vert s_j \Vert
+ \mathcal{O} \left( \Vert  \widehat{s} \Vert^{2} \right).
\end{equation*}
\end{theorem}
}

We can see that for cell matching using the encoder $\mathbf{\Phi}$ of DiffeoInvariantNet, the error between the test cell and the matched reference cell in latent space is bounded above by some functions of the minimal covering radius $\epsilon$ of the cell bank ${\widehat{G}}$ and the Lipschitz constant $L$ of the encoder.

\section{Empirical Results}
\subsection{Unit test on DiffeoInvariantNet}

First, we evaluated DiffeoInvariantNet's ability to match new cells with their most similar archetype in the cell bank. For quantitative comparison, we found the latent space distance between an augmented cell patch and its canonical counterpart compared to other cell patches. We then calculated the embedding fidelity measured by mean average precision~(MAP) as defined in~\cite{MAP_graph}, as well as the match accuracy for one or three neighbors. The results in Table~\ref{tab:cell_matching} show that DiffeoInvariantNet can organize a highly ordered latent space, allowing it to identify the correct cell matches.

\begin{table}[H]
\centering
\caption{Unit Test on DiffeoInvariantNet: Cell matching on histology images~\cite{MoNuSeg}.}
\label{tab:cell_matching}

\begin{tabular}{@{}lccccc@{}}
\toprule
 & MAP & 1-neighbor Accuracy \quad &3-neighbor Accuracy \quad \\
\midrule
Breast Cancer
& $0.954 \pm {\color{gray}{0.023}}$
& $0.949 \pm {\color{gray}{0.009}}$
& $0.912 \pm {\color{gray}{0.013}}$ \\
Colon Cancer
& $0.900 \pm {\color{gray}{0.004}}$
& $0.845 \pm {\color{gray}{0.006}}$
& $0.830 \pm {\color{gray}{0.007}}$ \\
Prostate Cancer
& $0.876 \pm {\color{gray}{0.012}}$ 
& $0.799 \pm {\color{gray}{0.055}}$
& $0.808 \pm {\color{gray}{0.015}}$ \\
\bottomrule
\end{tabular}
\end{table}

\subsection{Unit test on DiffeoMappingNet}

\begin{figure}[H]
    \centering
    \includegraphics[width=\textwidth]{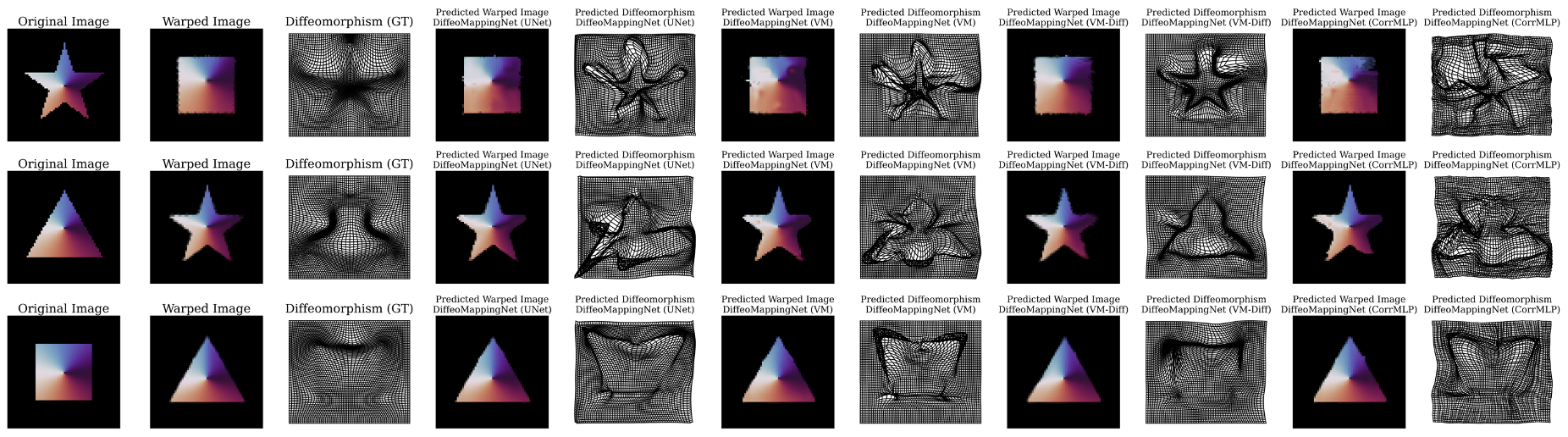}
    \caption{Mapping diffeomorphisms of synthetic shapes with DiffeoMappingNet. Qualitatively, VM-Diff~\cite{VM_diff} is the most competitive architecture for DiffeoMappingNet.}
    \label{fig:DiffeoMappingNet_results}
\end{figure}

We evaluated DiffeoMappingNet in isolation by predicting diffeomorphisms between pairs of synthetic shapes. Qualitatively, among the four architecture candidates, VM-Diff predicts diffeomorphisms most faithfully, as illustrated in Fig.~\ref{fig:DiffeoMappingNet_results}. Quantitatively, VM-Diff also emerges as the top candidate, especially when considering its $16 \times$ faster runtime compared to CorrMLP, which is otherwise equally capable~(see Table~\ref{tab:DiffeoMappingNet_results}). Hence, we selected VM-Diff as the architecture for DiffeoMappingNet in all the following experiments.

\begin{table}[H]
\centering
\caption{Unit Test on DiffeoMappingNet: Diffeomorphism prediction on synthetic shapes. Quantitatively, VM-Diff~\cite{VM_diff} is the most competitive architecture for DiffeoMappingNet.}
\begin{tabular}{lcccc}
\toprule
& UNet~\cite{UNet} & VM~\cite{VM} & VM-Diff~\cite{VM_diff} & CorrMLP~\cite{CorrMLP} \\
\midrule

$\textrm{NCC}~(\mathcal{W})$ $\uparrow$ & $\underline{-0.096} \pm {\color{gray} 0.961}$ & $-0.310 \pm {\color{gray} 0.899}$ & $\textbf{0.668} \pm {\color{gray} 5.397}$ & $-0.609 \pm {\color{gray} 0.527}$ \\
$D_\textrm{L1}~(\mathcal{W})$ $\downarrow$ & $1.758 \pm {\color{gray} 0.443}$ & $1.386 \pm {\color{gray} 0.232}$ & $\textbf{1.298} \pm {\color{gray} 0.258}$ & $\underline{1.356} \pm {\color{gray} 0.087}$ \\
$D_\textrm{L1}\textrm{~(image)}$ $\downarrow$ & $28.367 \pm {\color{gray} 2.937}$ & $27.180 \pm {\color{gray} 5.559}$ & $\textbf{26.621} \pm {\color{gray} 3.712}$ & $\underline{26.701} \pm {\color{gray} 3.675}$ \\
$\textrm{DSC}\textrm{~(mask)}$ $\uparrow$ & $0.964 \pm {\color{gray} 0.014}$ & $0.957 \pm {\color{gray} 0.020}$ & $\underline{0.966} \pm {\color{gray} 0.012}$ & $\textbf{0.972} \pm {\color{gray} 0.012}$ \\
$\textrm{IoU}\textrm{~(mask)}$ $\uparrow$ & $0.931 \pm {\color{gray} 0.025}$ & $0.918 \pm {\color{gray} 0.036}$ & $\underline{0.935} \pm {\color{gray} 0.023}$ & $\textbf{0.946} \pm {\color{gray} 0.022}$ \\
Runtime $\downarrow$ & $19.067 \pm {\color{gray} 1.424}$ & $\textbf{2.243} \pm {\color{gray} 0.130}$ & $\underline{3.220} \pm {\color{gray} 0.153}$ & $53.281 \pm {\color{gray} 1.602}$ \\

\bottomrule
\end{tabular}
\label{tab:DiffeoMappingNet_results}
\end{table}

\subsection{Application 1: Cell counting}
The first application we tackle is cell counting. This is done by raster scanning the entire microscopy image and assessing each image patch. These patches are encoded by DiffeoInvariantNet, and the embedding vectors are compared with those of the archetypal cells in the cell bank, as well as background image patches. A cell likelihood score is computed based on the distance in the latent space. Non-maximum suppression~\cite{NMS} is applied as postprocessing. DiffKillR significantly outperforms standard blob detection on cell counting.

\begin{table}[H]
\centering
\caption{Cell Counting Performance on histology images~\cite{MoNuSeg}.}

\begin{tabular}{lcccc}
\toprule
& & Precision~$\uparrow$ & Recall~$\uparrow$ & F1~$\uparrow$ \\
\midrule
Breast Cancer & Blob Detection & $0.488 \pm {\color{gray} 0.001}$ & $0.269 \pm {\color{gray} 0.020}$ & $0.347 \pm {\color{gray} 0.019}$ \\

\vspace{4pt} & DiffKillR~\textbf{(ours)}, 10\% 
& $\textbf{0.500} \pm {\color{gray} 0.076}$ 
& $\textbf{0.719} \pm {\color{gray} 0.003}$ 
& $\textbf{0.585} \pm {\color{gray} 0.054}$\\

Colon Cancer & Blob Detection 
& $0.323 \pm {\color{gray} 0.070}$
& $0.260 \pm {\color{gray} 0.044}$
& $0.288 \pm {\color{gray} 0.055}$ \\

\vspace{4pt} & DiffKillR~\textbf{(ours)}, 10\% 
& $\textbf{0.410} \pm {\color{gray} 0.051}$ 
& $\textbf{0.500} \pm {\color{gray} 0.053}$ 
& $\textbf{0.450} \pm {\color{gray} 0.051}$\\

Prostate Cancer & Blob Detection 
& $0.343 \pm {\color{gray}0.038}$
& $0.264 \pm {\color{gray}0.053}$ 
& $0.298 \pm { \color{gray}0.048}$ \\
 
& DiffKillR~\textbf{(ours)}, 10\% 
& $\textbf{0.464} \pm {\color{gray} 0.077}$ 
& $\textbf{0.640} \pm {\color{gray} 0.046}$ 
& $\textbf{0.531} \pm {\color{gray} 0.034}$  \\
\bottomrule
\end{tabular}
\label{tab:CellCounting_results}
\end{table}

\subsection{Application 2: Cell orientation prediction}

Next, we evaluated DiffKillR for predicting cell orientations of epithelial cells, where the apical-basal axes are manually annotated by a pathologist. To integrate these annotations into our framework, we converted the orientations into pixel-level gradient labels that indicate the annotated direction. We then applied the DiffKillR framework to this dataset. As shown in Table~\ref{tab:cell_orientation}, applying the annotation from the best match archetype yields an angular error of 30 degrees, and the full DiffKillR framework reduces it by 68\%.

\begin{table}[H]
\centering
\caption{Cell orientation prediction on epithelial cells, evaluated by L1 distance and angular difference in degrees. NCC: normalized cross correlation. MI: mutual information.}
\label{tab:cell_orientation}
\scalebox{0.85}{
\begin{tabular*}{1.13\linewidth}{@{\extracolsep{\fill}}lcccc}
\toprule
& Hard Example & Metric to Identify & \multirow{2}{*}{$D_\textrm{L1}\textrm{~(label)}$ $\downarrow$} & \multirow{2}{*}{$D_\theta\textrm{~(label)}$ $\downarrow$} \\
& Mining Ratio & Best Flip \& Rotation & \\
\midrule
Matching Archetype's Label & -- & -- & $0.246 \pm {\color{gray}0.036}$ & $30.29 \pm {\color{gray}4.57}$\\
\midrule
Flipping \& 90-degree rotations & -- & NCC & $0.207 \pm {\color{gray}0.025}$ & $19.67 \pm {\color{gray}7.22}$\\
DiffKillR~\textbf{(ours)}
 & 0.00 & NCC & $\underline{0.175} \pm {\color{gray}0.030}$ & $\underline{18.29} \pm {\color{gray}6.90}$ \\
 & 0.25 & NCC & $\textbf{0.168} \pm {\color{gray}0.025}$ & $\textbf{17.68} \pm {\color{gray}6.43}$\\
 & 0.50 & NCC & $0.189 \pm {\color{gray}0.028}$ & $19.01 \pm {\color{gray}7.25}$\\
 & 0.75 & NCC & $0.191 \pm {\color{gray}0.029}$ & $19.06 \pm {\color{gray}6.79}$\\
 & 1.00 & NCC & $0.187 \pm {\color{gray}0.076}$ & $19.54 \pm {\color{gray}7.21}$\\
\midrule
Flipping \& 90-degree rotations & -- & MI & $0.186 \pm {\color{gray}0.021}$ & $11.34 \pm {\color{gray}7.29}$\\
DiffKillR~\textbf{(ours)}
 & 0.00 & MI & $\underline{0.152} \pm {\color{gray}0.024}$ & $\underline{10.25} \pm {\color{gray}6.31}$ \\
 & 0.25 & MI & $\textbf{0.151} \pm {\color{gray}0.039}$ & ~~$\textbf{9.74} \pm {\color{gray}5.81}$\\
 & 0.50 & MI & $0.178 \pm {\color{gray}0.020}$ & $10.40 \pm {\color{gray}6.70}$\\
 & 0.75 & MI & $0.180 \pm {\color{gray}0.027}$ & $10.48 \pm {\color{gray}6.53}$\\
 & 1.00 & MI & $0.196 \pm {\color{gray}0.031}$ & $11.21 \pm {\color{gray}6.83}$\\
\bottomrule
\end{tabular*}
}
\end{table}

\subsection{Application 3: Few-shot cell segmentation}

\begin{figure}[H]
    \centering
    \includegraphics[width=\textwidth]{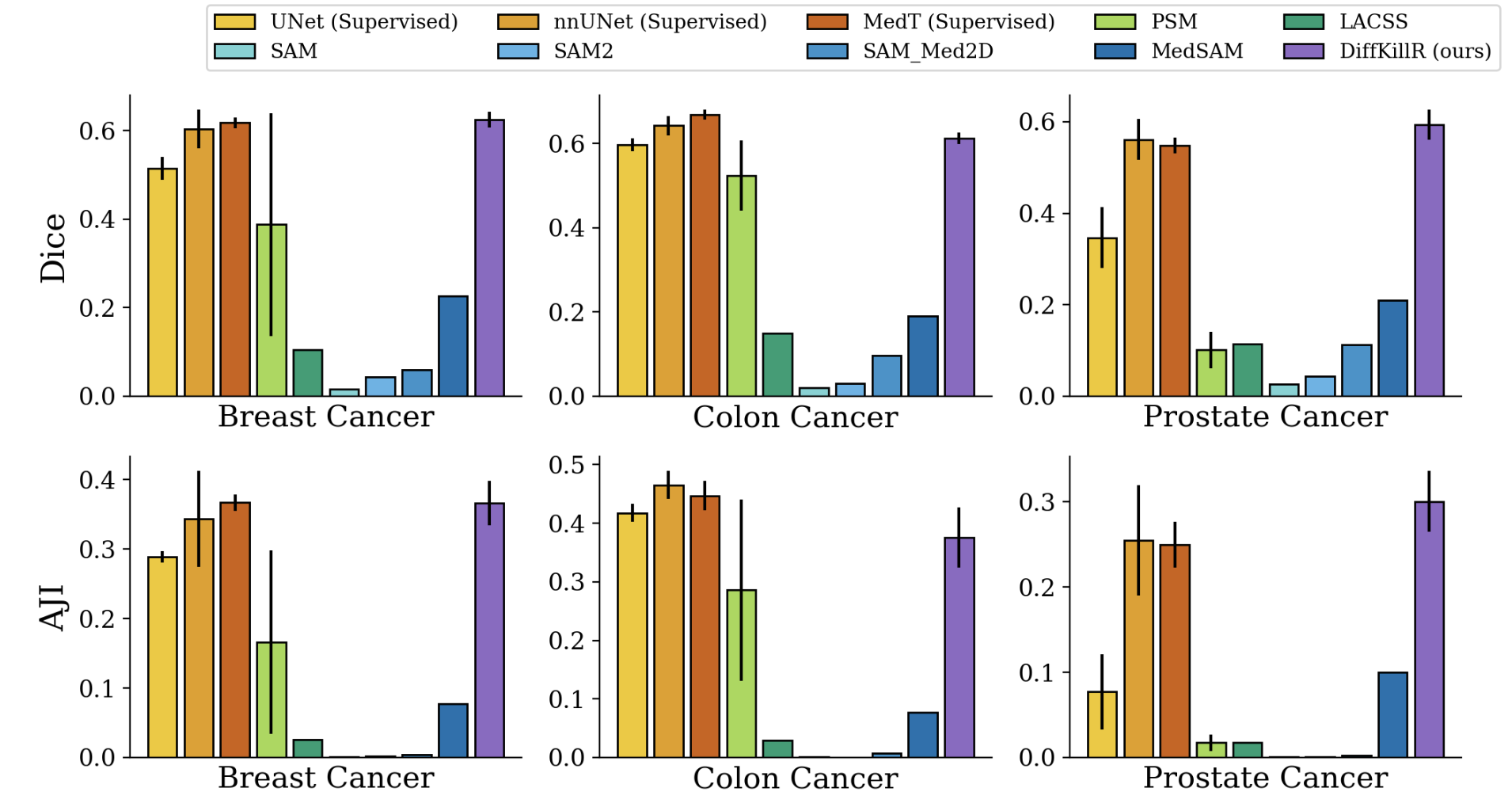}
    \vspace{-8pt}
    \caption{Few-shot cell segmentation performance on histology images~\cite{MoNuSeg}.}
    \label{fig:few_shot_segmentation}
    \vspace{-8pt}
\end{figure}

In our final case study, we investigated few-shot cell segmentation. We applied DiffKillR on microscopy images of 3 types of cancer cells, and benchmarked the few-shot segmentation performance against existing solutions. Baselines include methods that use supervised learning (UNet~\cite{UNet}, nn-UNet~\cite{nnUNet} and medical transformer~\cite{MedT}), unsupervised learning (LACSS~\cite{LACSS, LACSS2}), semi-supervised learning (PSM~\cite{PSM}), and Segment Anything models (SAM~\cite{SAM}, SAM2~\cite{SAM2}, SAM-Med2D~\cite{SAM_Med2D}, and MedSAM~\cite{MedSAM}). As shown in Figure~\ref{fig:few_shot_segmentation}, our proposed DiffKillR can achieve performance comparable to supervised methods using only 10\% of the training data, and it surpasses other unsupervised and semi-supervised methods, including the latest SAM-based models, by a large margin.

\section{Conclusion}
We propose DiffKillR, a dual-network framework that addresses the challenge of annotation scarcity in microscopy images through a novel approach of ``killing'' and ``recreating'' diffeomorphisms with a diffeomorphism-invariant network and a diffeomorphism-sensitive network. This innovative framework shows a promising direction to reduce the need for extensive manual labeling in various microscopy tasks.

\section{Acknowledgements}
This work was supported in part by the National Science Foundation (NSF Career Grant 2047856) and the National Institute of Health (NIH 1R01GM130847-01A1, NIH 1R01GM135929-01).

\clearpage
\newpage
\bibliographystyle{unsrt}
\bibliography{references}

\end{document}